\documentclass[conference]{IEEEtran}
\IEEEoverridecommandlockouts
\usepackage{cite}
\usepackage{amsmath,amssymb,amsfonts}
\usepackage{algorithmic}
\usepackage{graphicx}
\usepackage{textcomp}
\usepackage{xcolor}
\usepackage{booktabs}

\usepackage{algorithm}
\usepackage{algorithmic}

\def\BibTeX{{\rm B\kern-.05em{\sc i\kern-.025em b}\kern-.08em
    T\kern-.1667em\lower.7ex\hbox{E}\kern-.125emX}}

\begin{document}
\title{MetaRF: Differentiable Random Forest for Reaction Yield Prediction with a Few Trails\\
\thanks{This work is supported by XXX(Grant No. XXX).}
\thanks{*Correspondence author: Guangyong Chen (gychen@zhejianglab.com);}}
\author{\IEEEauthorblockN{Kexin Chen}
\IEEEauthorblockA{\textit{Department of Computer Science and Engineering} \\
\textit{The Chinese University of Hong Kong}\\
New Territories, Hong Kong SAR \\
kxchen@cse.cuhk.edu.hk}
\and
\IEEEauthorblockN{Guangyong Chen$^*$}
\IEEEauthorblockA{\textit{Zhejiang Lab} \\
\textit{Zhejiang University}\\
Hangzhou, China \\
gychen@zhejianglab.com}
\and
\IEEEauthorblockN{Junyou Li}
\IEEEauthorblockA{\textit{Zhejiang Lab} \\
Hangzhou, China \\
\text{ }lijunyou@zhejianglab.com}
\and
\IEEEauthorblockN{\text{ }\text{ }\text{ }\text{ }\text{ }\text{ }\text{ }\text{ }\text{ }\text{ }\text{ }\text{ }\text{ }\text{ }\text{ }\text{ }\text{ }\text{ }\text{ }\text{ }\text{ }Yuansheng Huang}
\IEEEauthorblockA{\text{ }\text{ }\text{ }\text{ }\text{ }\text{ }\text{ }\text{ }\text{ }\text{ }\text{ }\text{ }\text{ }\text{ }\text{ }\text{ }\text{ }\text{ }\text{ }\text{ }\text{ }\text{ }\text{ }\text{ }\text{ }\textit{Zhejiang Lab} \\
\text{ }\text{ }\text{ }\text{ }\text{ }\text{ }\text{ }\text{ }\text{ }\text{ }\text{ }\text{ }\text{ }\text{ }\text{ }\text{ }\text{ }\text{ }\text{ }\text{ }\text{ }\text{ }\text{ }\text{ }\text{ }\textit{Zhejiang University}\\
\text{ }\text{ }\text{ }\text{ }\text{ }\text{ }\text{ }\text{ }\text{ }\text{ }\text{ }\text{ }\text{ }\text{ }\text{ }\text{ }\text{ }\text{ }\text{ }\text{ }\text{ }\text{ }\text{ }\text{ }\text{ }Hangzhou, China \\
\text{ }\text{ }\text{ }\text{ }\text{ }\text{ }\text{ }\text{ }\text{ }\text{ }\text{ }\text{ }\text{ }\text{ }\text{ }\text{ }\text{ }\text{ }\text{ }\text{ }\text{ }\text{ }\text{ }\text{ }\text{ }22219109@zju.edu.cn}
\and
\IEEEauthorblockN{Pheng-Ann Heng}
\IEEEauthorblockA{\textit{Department of Computer Science and Engineering} \\
\textit{The Chinese University of Hong Kong}\\
New Territories, Hong Kong SAR \\
pheng@cse.cuhk.edu.hk}
}
\maketitle

\begin{abstract}
Artificial intelligence has deeply revolutionized the field of medicinal chemistry with many impressive applications, but the success of these applications requires a massive amount of training samples with high-quality annotations, which seriously limits the wide usage of data-driven methods. In this paper, we focus on the reaction yield prediction problem, which assists chemists in selecting high-yield reactions in a new chemical space only with a few experimental trials. To attack this challenge, we first put forth MetaRF, an attention-based differentiable random forest model specially designed for the few-shot yield prediction, where the attention weight of a random forest is automatically optimized by the meta-learning framework and can be quickly adapted to predict the performance of new reagents while given a few additional samples. To improve the few-shot learning performance, we further introduce a dimension-reduction based sampling method to determine valuable samples to be experimentally tested and then learned. Our methodology is evaluated on three different datasets and acquires satisfactory performance on few-shot prediction. In high-throughput experimentation (HTE) datasets, the average yield of our methodology's top 10 high-yield reactions is relatively close to the results of ideal yield selection.
\end{abstract}


\section{Introduction}

Computer-aided synthesis planning (CASP)\cite{b43}, which aims to assist chemists in synthesizing new molecule compounds, has been rapidly transformed by artificial intelligence methods. Given the availability of large-scale reaction datasets, such as the United States Patent and Trademark Office (USPTO)\cite{b42}, Reaxys\cite{b44}, and SciFinder\cite{b45}, CASP has become an increasingly popular topic in pharmaceutical discovery and organic chemistry with many impressive breakthroughs achieved\cite{b3}. The current CASP systems can be divided into two critical aspects, retrosynthetic planning and forward-reaction prediction\cite{b36}. Retrosynthetic planning, including template-based and template-free methods, can help generate possible synthetic routes of target molecules\cite{b37}. Forward-reaction prediction is mainly used to evaluate the strategies proposed by retrosynthetic planning and increase the likelihood of experimental success\cite{b38}. However, without considering reaction yield or reaction conditions, the synthetic strategies proposed in the CASP systems would be difficult to be implemented. It still remains a big challenge to predict the reaction yield. Due to the complexity of chemical experiments, few solid theories can help predict the reaction yield of a new chemical reaction given a specific condition, let alone optimize a reaction condition, which heavily depends on expertise, knowledge, intuition, numerous practices, extensive literature reading and even the luck of chemists\cite{b3,b32}.


Some pioneer efforts have been contributed to predict the reaction yield, and then find the optimal reaction condition. Note that the optimal reaction selection problem can be naturally treated as a classical out-of-distribution (OOD) problem, since the optimal reaction is often not included in the training set. Ahneman et al.\cite{b4} reported that the random forest model achieved the best performance on OOD yield prediction due to its good generalization ability. Zuranski et al.\cite{b21} reviewed and examined the OOD performance of different machine learning algorithms and reaction embedding techniques. Dong et al.\cite{b26} used the XGBoost model and achieved satisfactory OOD performance. Zhu et al.\cite{b27} demonstrated that regression-based machine learning had great application potential in OOD yield prediction. However, in OOD yield prediction, the relatively large difference between training and testing data deteriorates the predicting performance of the model. 

In this paper, we follow a more relaxed but practical setting, where we are allowed to add a few data of new reagents or conditions into the training set. Considering the limited amount of reaction condition data, few-shot yield prediction has great potential in solving this problem. Few-shot yield prediction adds very few reaction samples(e.g. around five samples) from new reagents or conditions into training data. It is reasonable to hypothesize that using data of a new reagent can improve prediction results. Questions yet to be explored are how to use these new samples, which sample to select, and how much data from the new reagent leads to a satisfactory result.

To bridge this gap, we proposed MetaRF, an attention-based differentiable random forest model with a meta-learning technique applied to determine attention weights adaptively. The random forest has been proved as an ensemble method with outstanding performance on datasets with small sample size\cite{b23,b24}.

However, the structure of random forest is non-differential, which is hard to combine with the gradient-based techniques in meta-learning. To solve this problem and achieve robust performance on new reagents, we propose to add attention weights to the random forest through a meta-learning framework, Model Agnostic Meta-Learning (MAML) algorithm\cite{b19}. The key idea of MAML is to train the model's initial parameters so that the model can quickly adapt to a new task after the parameters have been updated through a few gradient steps computed with few-shot data from that new task\cite{b19}. MAML is applied to determine the attention weights of decision trees in the random forest so that the model can quickly adapt to predict the performance of new reagents using few-shot training samples.
The choice of few-shot training samples also has a significant influence on model performance. Few-shot learning can have better-predicting performance if it is allowed to choose the training samples\cite{b40}. To tackle this challenge, we use Kennard-Stone (KS) algorithm\cite{b14} to select the most representative samples which cover the experimental space homogeneously. Since the KS algorithm is based on Euclidean distance, which suffers from the curse of dimensionality\cite{b41}, T-distributed stochastic neighbor embedding (TSNE)\cite{b8} is applied for unsupervised nonlinear dimension reduction.

Our methodology is comprehensively evaluated on Buchwald-Hartwig high-throughput experimentation (HTE) dataset\cite{b4}, Buchwald-Hartwig electronic laboratory notebooks (ELN) dataset\cite{b25} and Suzuki–Miyaura HTE dataset\cite{b7}. In Buchwald–Hartwig high-throughput experimentation (HTE) dataset, our method achieves $R^{2}$=0.648 using 2.5\% of the dataset as the training set. To reach a comparable result, the baseline method (random forest) needs to use at least 20\% of the dataset as the training set. With the help of 5 additional samples, our method can effectively explore unseen chemical space and select high-yield reactions. The 10 reactions, which are predicted to have the highest yield, reach an average yield of 93.7\%, relatively close to the result of ideal yield selection (95.5\%). In contrast, the top 10 high-yield reactions selected by the baseline method have an average yield of 86.3\%, and the average yield of random selection is 52.1\%.

\begin{figure*}[tbp]
\centerline{\includegraphics[width=\textwidth]{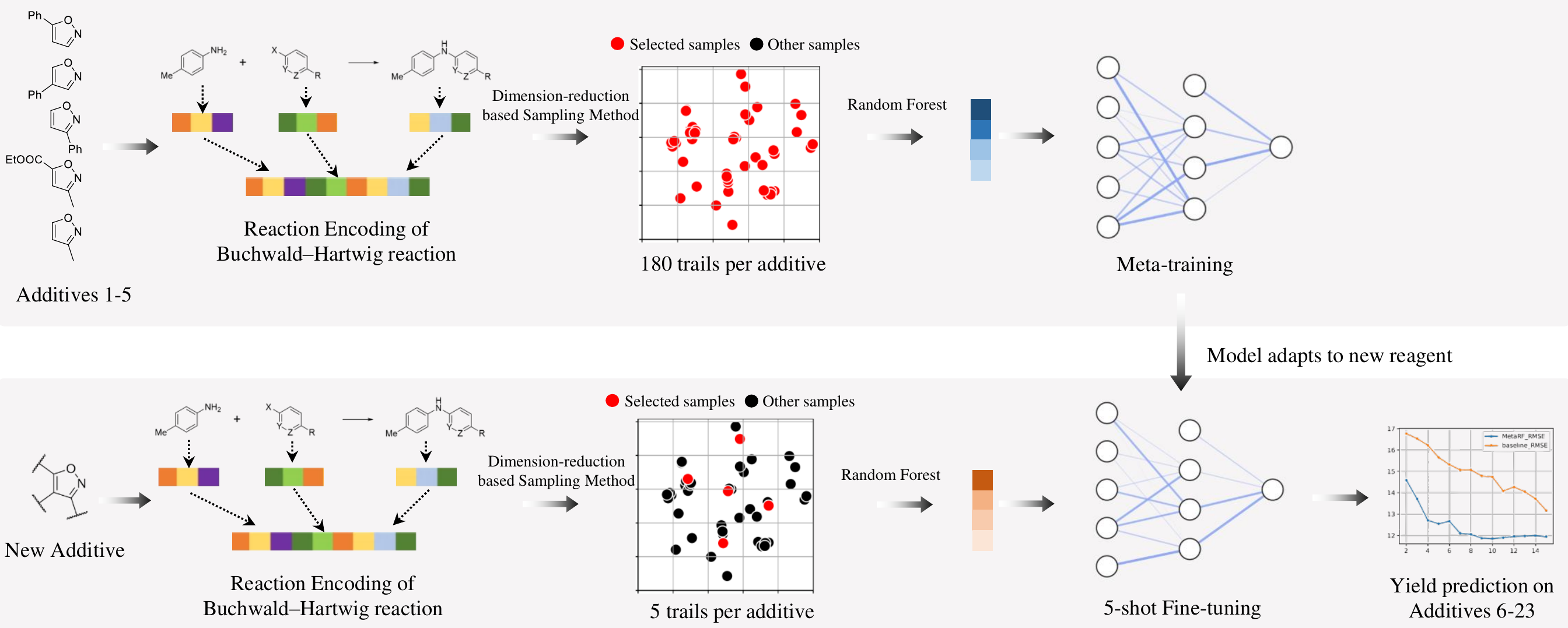}}
\caption{Workflow of this research that includes reaction encoding, dimension-reduction based sampling method, and attention-based differentiable random forest model. Buchwald–Hartwig HTE dataset is taken as an example.}
\label{overview}
\end{figure*}

The overview framework of this research is presented in Fig.~\ref{overview}. More details of methodology are in Section \ref{method}.
The methodology in this paper can predict the effect of a new reagent structure with few reaction data, and our sampling method can help chemists choose the order of experiments.

\section{Methods}\label{method}

\subsection{Reaction Encoded with DFT}
Density Functional Theory (DFT) descriptor is widely used in molecular embedding owing to its strong and effective feature generation ability\cite{b6}. Previous research\cite{b21} shows that the DFT descriptor provides transferable chemical insight and sheds light on the underlying mechanism. We followed the DFT descriptor calculation in \cite{b4}, which includes molecular, atomic, and vibrational property descriptors. As in \cite{b4}, we generate the numerical encoding of each reaction by concatenating the DFT descriptor of each chemical component. 

For example, the encoding of experiment $i$ in Buchwald–Hartwig reaction is
\begin{equation}
{{{x}}}_{i}={{{x}}}_{{\rm{Aryl\text{ }halide}}}\oplus {{{x}}}_{{\rm{Pd\text{ }catalyst}}}\oplus {{{x}}}_{{\rm{Additive}}}\oplus {{{x}}}_{{\rm{Base}}}
\end{equation}
where $\oplus$ denotes concatenation and ${{{x}}}_{{\rm{Aryl\text{ } halide}}}$, ${{{x}}}_{{\rm{Pd\text{ } catalyst}}}$, ${{{x}}}_{{\rm{Additive}}}$, ${{{x}}}_{{\rm{Base}}}$ denotes DFT descriptor vector of the corresponding Aryl halide, Pd catalyst, Additive and Base.

\subsection{MetaRF: Attention-based Differentiable Random Forest}  

\subsubsection{Random Forest}\label{subsubrf}
The random forest model is an extension of bagging methodology\cite{b23}. This algorithm is based on a collection of decision trees, especially classification and regression trees (CART). Each decision tree is learned independently on a group of bootstrapped samples. Random forest is chosen as the base model for its outstanding performance in small datasets. To explore the OOD predicting ability, the testing set and validation set must include at least one unseen reagent in the training set. For example, among the 22 different additives in Buchwald–Hartwig HTE dataset, 4 additives are used for training, 1 additive is used for validation, and 17 for testing. In this way, the training set and validation set take 22.7\% of the dataset. To further reduce the size of the training set, we use the sampling method in Section \ref{subtsne}. The random forest model is trained on the reduced training set. 

In the random forest model, forest $\mathcal{F}$ is a collection of decision trees:

\begin{equation}
\mathcal{F}(\Theta )=\{{{h}_{m}}(\mathbf{x};{{\Theta }_{m}})\},m=1,2,\ldots M
\end{equation}
where $M$ is the total number of decision trees, $\Theta =\{{{\Theta }_{1}},{{\Theta }_{2}},\ldots {{\Theta }_{M}}\}$ represents parameters in $\mathcal{F}$, which includes splitting variables and their splitting values. $\mathcal{F}$ is fitted by the training data $\mathcal{L}=\left\{ ({{x}_{1}},{{y}_{1}}),\cdot \cdot \cdot ({{x}_{N}},{{y}_{N}}) \right\}$, where ${{x}_{i}}$ is the embedding of reaction $i$ (defined in the former section) and ${{y}_{i}}$ represents the yield of the reaction.

The decision tree is a simple predictive model. It has the form

\begin{equation}
{{h}_{m}}(x)=\sum\nolimits_{j=1}^{J}{{{b}_{jm}}}I(x\in {{R}_{jm}})
\end{equation}
where $J$ is the number of its leaves. The tree partitions the input space into $J$ disjoint regions ${{R}_{1m}},\ldots,{{R}_{Jm}}$ and predicts a constant value in each region. ${{b}_{jm}}$ is the value predicted in ${{R}_{jm}}$.

At each tree node, part of the variables are randomly selected as a subset. The splitting variable is chosen from this subset. This random selection of features at each node decreases the correlation between the trees in the forest and thus reduces the error rate of the random forest.

Concating the results of each decision tree ${{h}_{m}}(x)$, we have

\begin{equation}
{{x}_{i}}^{\prime }=\left[ \begin{matrix}
   \begin{matrix}
   \begin{matrix}
   {{h}_{1}}({{x}_{i}})  \\
\end{matrix}  \\
   {{h}_{2}}({{x}_{i}})  \\
   \vdots   \\
\end{matrix}  \\
   {{h}_{M}}({{x}_{i}})  \\
\end{matrix} \right]
\end{equation}

\subsubsection{Differentiable RF Learned with Meta Learning}

Although the random forest is a robust algorithm in yield prediction, it remains a challenge to combine random forest with few-shot learning techniques in yield prediction of new reagents or conditions. Meta-learning introduces a model that can quickly adapt to new tasks with few additional samples. Model Agnostic Meta-Learning (MAML) framework\cite{b19} is a well-known meta-learning approach with both simplicity and effectiveness. However, the non-differential characteristic of the random forest makes it difficult to integrate with the gradient-based meta-learning framework. To tackle this problem, we solve different attention weights to decision trees in the random forest using MAML framework, which consists of a meta-training phase and a few-shot fine-tuning phase.

In the meta-training phase, MAML provides a good initialization of parameters in deep networks. Assume $\theta$ is the parameters that need to be optimized and ${f}_{\theta}$ is the parametrized function. In each training iteration, the updated $\theta$ is computed using one gradient descent update on task ${T}_{i}$, and the loss function is computed using the updated $\theta$. Sampling task ${T}_{i}$ includes two steps. An additive $S_i$ is randomly sampled from the training additive set. Then $K$ reactions with additive $S_i$ are randomly sampled to form task ${T}_{i}$. More concretely, the loss function is defined as follows:
\begin{equation}
\underset{\theta }{\mathop{\text{min}}}\,\text{  }\sum\limits_{{{T}_{i}}\sim T}{{{L}_{{{T}_{i}}}}({{f}_{\theta -\alpha  {{\nabla }_{\theta }} {{L}_{{{T}_{i}}}}({{f}_{\theta }})}})}
\end{equation}
where $L$ is the mean square error between the prediction ${f}_{\phi }({{x}_{j}}^{\prime })$ and true value ${y}_{j}$ in task ${T}_{i}$. 
\begin{equation}
{{L}_{{{T}_{i}}}}({{f}_{\phi }})=\sum\limits_{({{x}_{j}}^{\prime },{{y}_{j}})\sim {{T}_{i}}}{\left\| {{f}_{\phi }}({{x}_{j}}^{\prime })-{{y}_{j}} \right\|_{2}^{2}}
\end{equation}
${{x}_{j}}^{\prime }$ is computed using the method in Section \ref{subsubrf}. As in Finn et al.\cite{b19}, the regressor ${f}_{\theta}$ is a neural network with 2 hidden layers of size 40 with ReLU nonlinearities. During training, Equation (5) is minimized using gradient descent algorithm Adam\cite{b22} to acquire the parameter $\theta _{{\mathrm{meta-training}}}$. 

In the few-shot fine-tuning phase, the model is fine-tuned with a few samples from each testing additive. One iteration of gradient descent is performed to achieve $\theta _{{\mathrm{few - shot}}}$ suitable for the new task ${T}_{test}$:

\begin{equation}
\theta _{{\mathrm{few - shot}}} = \theta _{{\mathrm{meta-training}}} -\alpha  {{\nabla }_{\theta _{{\mathrm{meta-training}}}}} {{L}_{{{T}_{test}}}}({{f}_{\theta _{{\mathrm{meta-training}}}}})
\end{equation}

For each additive in the testing set, the fine-tune sample in ${T}_{test}$ is selected using dimension-reduction based sampling method in Section \ref{subtsne}. The number of fine-tune samples is altered in our experiments.

\subsection{Dimension-reduction based Sampling Method}\label{subtsne}

For the few-shot learning problem, the few-shot training samples have a significant influence on the training performance. If we preferentially select the most representative samples as training samples, the performance of few-shot learning can be dramatically improved\cite{b46}. We use Kennard-Stone (KS) algorithm\cite{b14} to select the most representative samples by selecting a new sample that has relatively large distances from previously selected samples. However, the KS algorithm uses Euclidean distance to represent the distances between samples, which is less effective in the high dimensional reaction data\cite{b47}. Thus we propose to add T-distributed stochastic neighbor embedding (TSNE)\cite{b8} before the KS algorithm to reduce the dimension of reaction data. TSNE is a widely used unsupervised nonlinear dimension reduction technique owing to its advantage in capturing local data characteristics and revealing subtle data structures\cite{b8,b9,b13}. From a chemical perspective, our dimension-reduction based sampling method can select representative samples with very different chemical structures and properties, which may shed light on the design of chemical experiments.

\subsubsection{T-distributed stochastic neighbor embedding (TSNE)}
TSNE is a nonlinear and unsupervised method that maps the high dimension data to low dimension, while retaining the significant structure of the original data\cite{b8,b10}. It is based on probabilistic modeling of data points in the original space and the projection space\cite{b11}.

The TSNE algorithm is based on the SNE framework\cite{b12}, which converts high dimensional Euclidean distances into conditional probabilities, representing similarities for every data pair. Given a set of high dimensional reaction embedding data ${{x}_{1}},{{x}_{2}},\ldots ,{{x}_{N}}$, the similarity of data ${{x}_{i}}$ to data ${{x}_{j}}$ is represented by the conditional probability $p_{i\mid j}$, 

\begin{equation}
p_{i\mid j}={\exp(-\Vert {\bf x}_{i}-{\bf x}_{j}\Vert^{2}/2\sigma_{i}^{2})\over {\sum}_{k\neq l}\exp(-\Vert {\bf x}_{k}-{\bf x}_{l}\Vert^{2}/2\sigma_{i}^{2})}
\end{equation}
where the parameter $\sigma_{i}$ is determined based on the dataset and this algorithm is not very sensitive to this parameter. For computational convenience, symmetric SNE defined joint probabilities as
\begin{equation}
p_{ij}={p_{i\mid j}+p_{j\mid i}\over {2N}}
\end{equation}

Assume each data point ${{x}_{i}}$ is mapped to a projection point ${{z}_{i}}\in \mathbb{R}^{d}$, where typically $d=2$. Unlike the probabilities in the original space, the Student's t-distribution is used in the projection space. Using this distribution, the joint probabilities $q_{ij}$ are defined as
\begin{equation}
q_{ij}={(1+\Vert {\bf z}_{i}-{\bf z}_{j}\Vert^{2})^{-1}\over {\sum}_{k\neq l}(1+\Vert {\bf z}_{k}-{\bf z}_{l}\Vert^{2})^{-1}}
\end{equation}

T-SNE aims at finding ${{z}_{i}}$ that minimizes the difference between these two probabilities, which is measured by the Kullback–Leibler divergence:
\begin{equation}
{\rm KL}(P\Vert Q)=\sum_{i}\sum_{j}p_{ij}\log\left({p_{ij}\over q_{ij}}\right)
\end{equation}
Typically the gradient descent technique is used for optimization.

\subsubsection{Kennard-Stone (KS) algorithm }
Kennard-Stone (KS) algorithm is a well-known method to select the most representative samples from the whole dataset\cite{b14,b16,b17}. The algorithm aims at choosing a subset of samples that cover the experimental space homogeneously\cite{b18}. First, the Euclidean distance between each pair of samples is calculated, and a pair of samples with the largest distance is chosen. Then the following samples are selected sequentially based on the distances to the already selected samples. The remaining sample with the largest distances is chosen and added to the subset. This procedure is repeated until a certain number of samples are selected.

\begin{figure}[tbp]
\centerline{\includegraphics[width=0.5\textwidth]{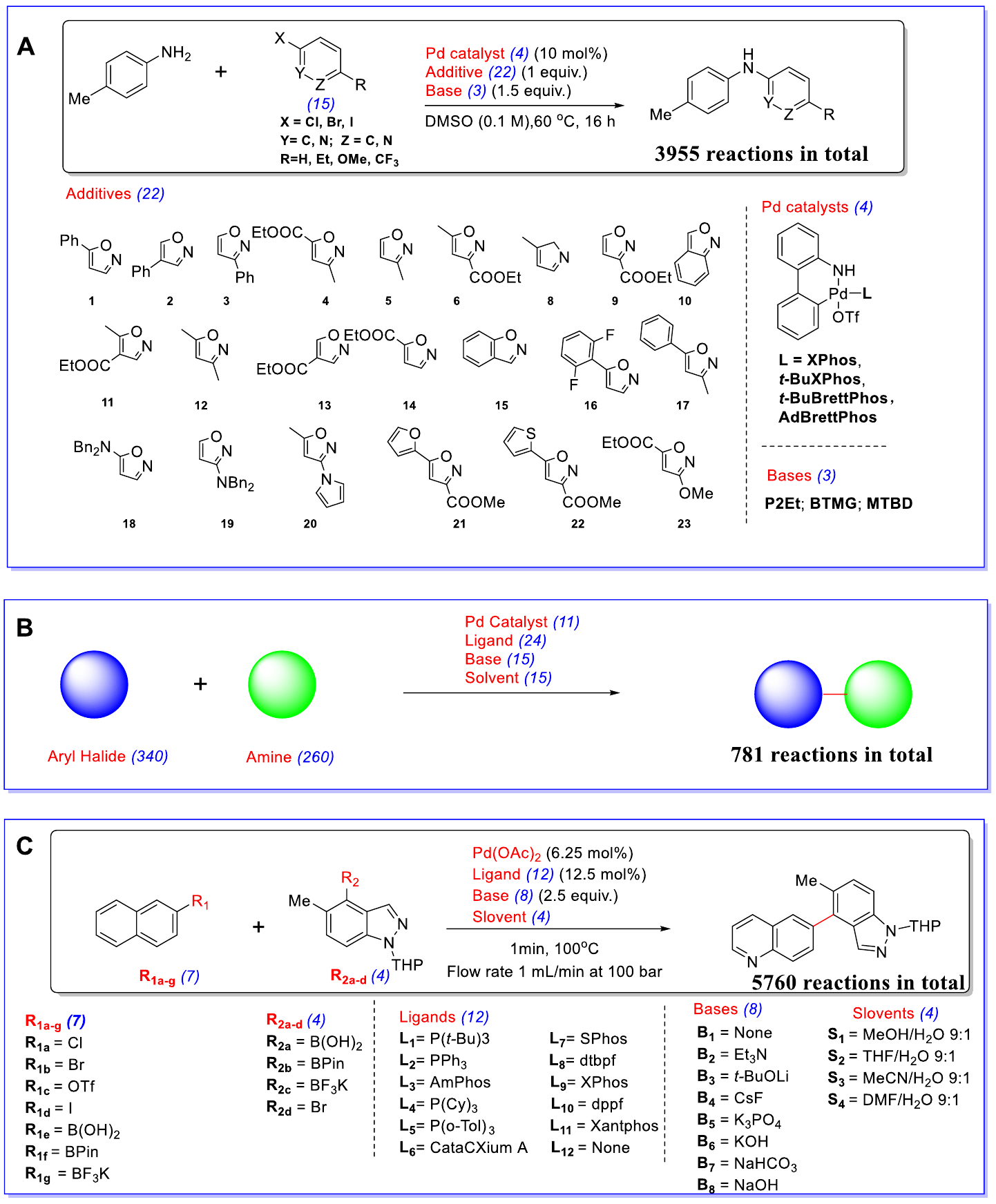}}
\caption{A) Buchwald–Hartwig HTE dataset. B) Buchwald–Hartwig ELN dataset. C) Suzuki–Miyaura HTE dataset. }
\label{dataset}
\end{figure}

\section{Results}\label{sec2}
\subsection{Performance Benchmarking}
\begin{figure*}[htbp]
\centerline{\includegraphics[width=0.95\textwidth]{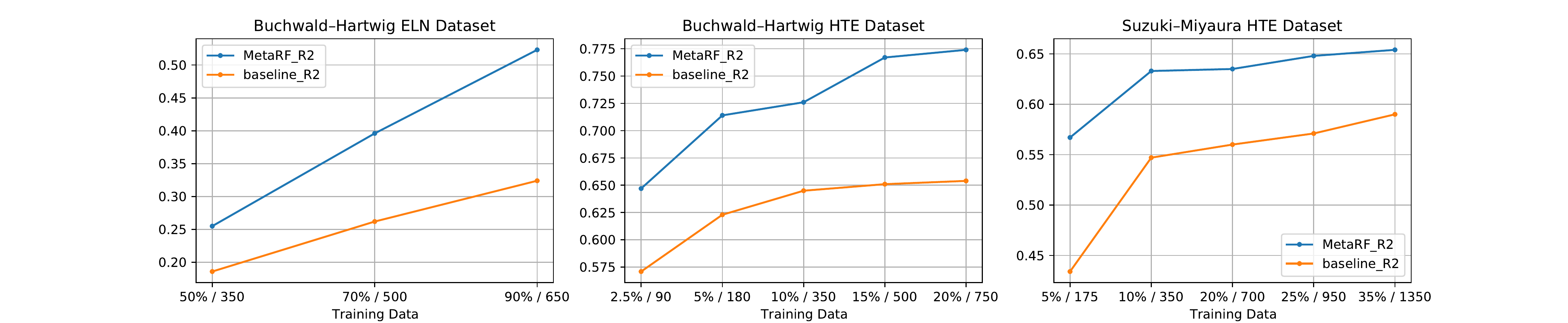}}
\caption{Comparison of test set performance of MetaRF and baseline on three datasets. $R^{2}$ performance increases gradually as the size of training data increases. MetaRF outperforms the baseline with markedly fewer training samples.}
\label{training_percentage}
\end{figure*}

\begin{figure*}[htbp]
\centerline{\includegraphics[width=0.95\textwidth]{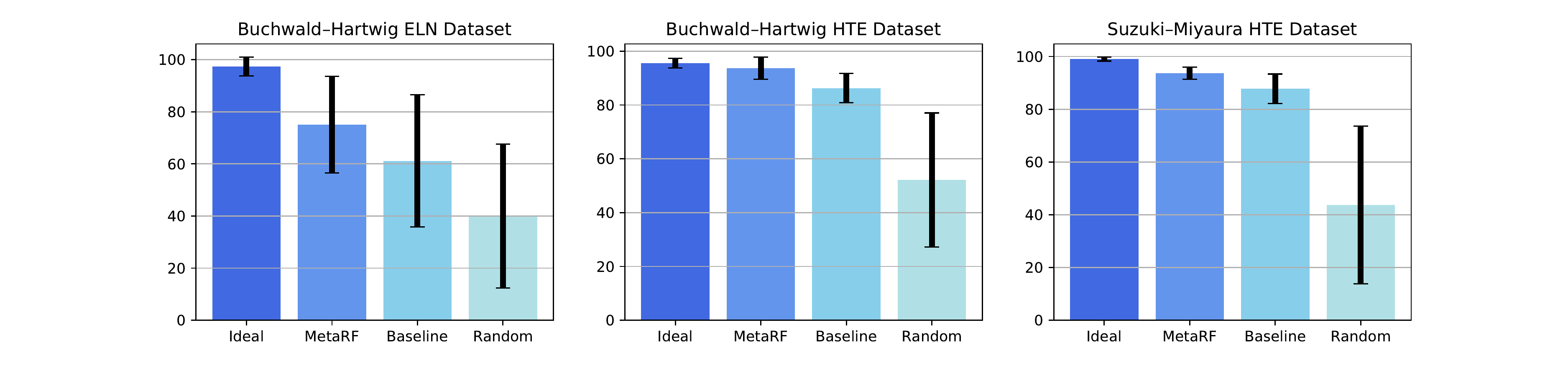}}
\caption{Average and standard deviation of the yield for the top 10 reactions predicted to have the highest yields.}
\label{high-yield}
\end{figure*}
We evaluate our method with Buchwald–Hartwig electronic laboratory notebooks (ELN) dataset\cite{b25}, Buchwald–Hartwig HTE dataset\cite{b4} and Suzuki–Miyaura HTE dataset\cite{b7}. Buchwald–Hartwig HTE dataset is the HTE results of the Pd-catalysed Buchwald-Hartwig cross-coupling reaction. This dataset consists of 3955 reactions as shown in Fig. \ref{dataset}A, and the reaction space is the combination of 15 aryl halides, 4 Buchwald ligands, 3 bases, and 22 isoxazole additives. Buchwald-Hartwig ELN dataset disclosed a real-world dataset from electronic laboratory notebooks (ELN) at AstraZeneca. The dataset covers a large reaction space. 340 aryl halides, 260 amines, 24 ligands, 15 bases, and 15 solvents should have covered 4.7 x 108 possible combinations. While in fact, the dataset only includes 781 reactions in Fig. \ref{dataset}B, resulting in a rather sparse coverage. HTE dataset greatly differs from the ELN dataset in the coverage of chemical space and characteristics. HTE dataset covers the entire search space of reaction condition while ELN dataset has a sparse coverage of wider chemical space. We also evaluate our methodology on Suzuki–Miyaura HTE dataset\cite{b7} to show that our methodology can be easily adapted to other reactions. Suzuki–Miyaura reaction means that aryl halide reacts with an organoboron compound to form a new C-C bond in the presence of Pd catalyst, ligand, and base. The mechanism of the Suzuki-Miyaura reaction is close to the Buchwald-Hartwig reaction, they all include an oxidative addition step and reductive elimination step in the catalytic cycle mechanism. The dataset includes 15 pairs of electrophiles and nucleophiles($R_{1a-d}$ with $R_{2a-c}$ and $R_{1e-g}$ with $R_{2d}$, 12 ligands, 8 bases, and 4 solvents, resulting in 5760 reactions in Fig. \ref{dataset}C. Experiments show that our methodology possesses outstanding performance on few-shot yield prediction.  

As shown in Fig.\ref{training_percentage}, our model outperforms the baseline method (random forest) on all three datasets when the size of training data increases gradually. For a fair comparison, we enlarge the training set of the baseline method with the additional fine-tune samples to guarantee that our method shares the same quantity of training data as the baseline method. Experiments show that our method possesses enhanced predictive power with markedly fewer training samples. For example, when trained on only 2.5\% of Buchwald–Hartwig HTE data, MetaRF acquires comparable results with the baseline method using 20\% of the same reaction data. 2.5\% of the Buchwald–Hartwig HTE data includes only 90 reactions. Using 2.5\% of the data as the training set, our method reaches $R^{2}$=0.648 while the $R^{2}$ of the baseline method is 0.571. When the training set increases to 20\% of the data, the $R^{2}$ of the baseline method is only 0.654. This comparison is similar to the results on Buchwald–Hartwig ELN and Suzuki–Miyaura HTE datasets. These results indicate that our method has great application potential in few-shot yield prediction. In our experiments, 80 training iterations are performed, and we use one gradient update with $K = 40$ examples and learning rate $\alpha= 0.0001$.
More details about the splitting of the training set, validation set, and testing set are in Section \ref{subsubrf}.

\begin{figure}[tbp]
\centerline{\includegraphics[width=0.5\textwidth]{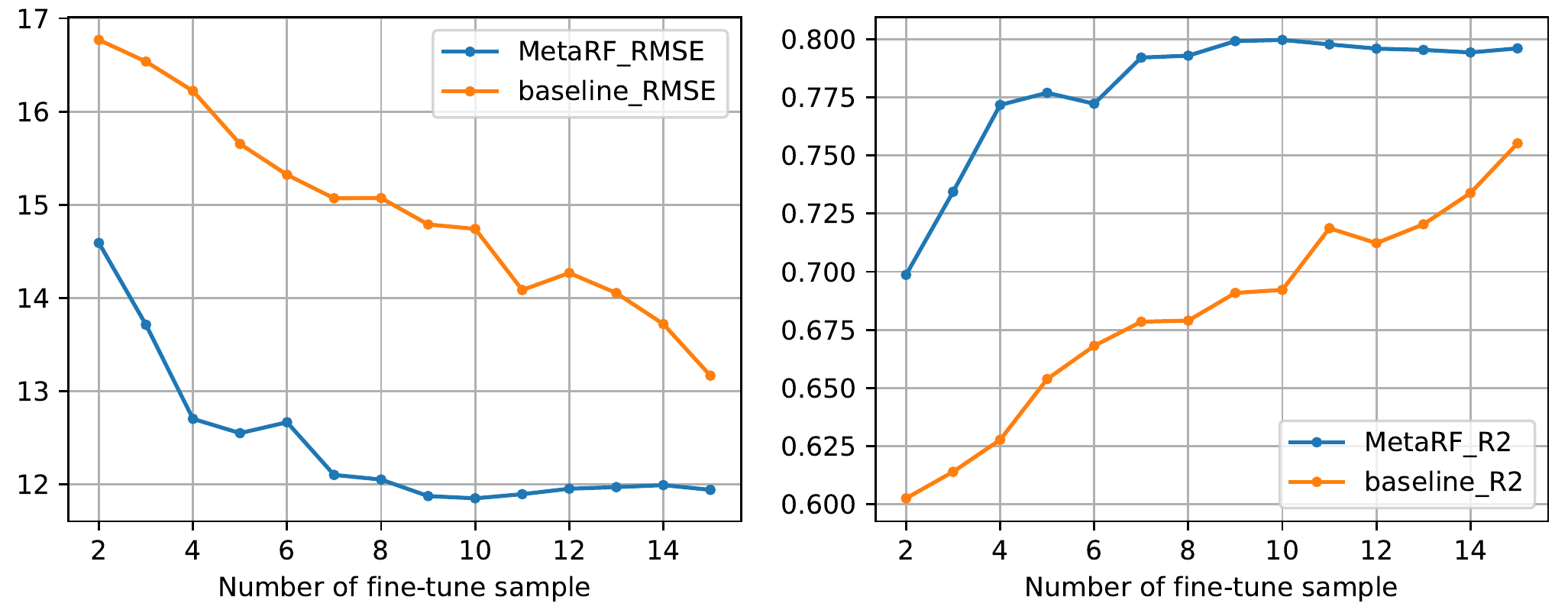}}
\caption{The RMSE and $R^{2}$ results when the number of
fine-tune sample increases gradually in the Buchwald–Hartwig HTE dataset.}
\label{finetune_sample}
\end{figure}

\begin{table}[ht]
\begin{center}
\begin{minipage}{240pt}
\caption{The results of ablation study(in Buchwald–Hartwig HTE dataset).}\label{ablation}%
\begin{tabular}{@{}lcc@{}}
\toprule 
& RMSE $\downarrow$  & $R^{2}$ $\uparrow$ \\
\midrule
Baseline    & 15.6535   & 0.6538  \\
\midrule
MetaRF + Dimension-reduction (proposed method)    & \textbf{12.6401}   & \textbf{0.7738}  \\
MetaRF + Random$^{\mathit{a}}$  & 14.5454  & 0.7003   \\
\midrule
MAML + Dimension-reduction$^{\mathit{b}}$   & 21.0006   & 0.3730 \\
MAML + Random    & 21.0204   & 0.3753  \\
\midrule
Transfer learning + Dimension-reduction$^{\mathit{c}}$   & 14.4575   & 0.7038   \\
Transfer learning + Random    & 15.2232   & 0.6705 \\
\bottomrule
\end{tabular}
\footnotetext[1]{The first ablation test, replacing the dimension-reduction based sampling with random sampling.}
\footnotetext[2]{The second ablation test, removing the random forest structure.}
\footnotetext[3]{The third ablation test, replacing MAML with transfer learning framework.}
\end{minipage}
\end{center}
\end{table}

\begin{table*}[ht]
\begin{center}
\begin{minipage}{.8\textwidth}
\caption{The relative improvement of MetaRF compared to the baseline method.(in Buchwald HTE dataset)}\label{margin}
\begin{tabular*}{\textwidth}{@{\extracolsep{\fill}}ccccccc@{\extracolsep{\fill}}}
\toprule%
& \multicolumn{3}{@{}c@{}}{RMSE $\downarrow$} & \multicolumn{3}{@{}c@{}}{$R^{2}$ $\uparrow$} \\\cmidrule{2-4}\cmidrule{5-7}%
Sample$^{\mathit{a}}$ & MetaRF & Baseline & Margin$^{\mathit{b}}$ & MetaRF & Baseline & Margin$^{\mathit{b}}$ \\
\midrule
2  & 14.5912 & 16.7692 & 12.99\% & 0.6986 & 0.6026 & 15.94\%\\
4  & 12.7039 & 16.2212  & 21.68\%  & 0.7717 & 0.6277 & 22.93\%\\
6  & 12.6674 & 15.3222  & 17.33\%  & 0.7722 & 0.6681 & 15.58\%\\
8  & 12.0527 & 15.0728 & 20.04\% & 0.7929 & 0.6790 & 16.77\%\\
10  & 11.8515 & 14.7419  & 19.61\%  & 0.7996 & 0.6922 & 15.52\%\\
\bottomrule
\end{tabular*}
\footnotetext[1]{The number of fine-tune samples.}
\footnotetext[2]{Relative improvement compared to the baseline method, random forest.}
\end{minipage}
\end{center}
\end{table*}

Then we test our method on the ability to search for reactions with the highest yield. This ability is valuable because it helps chemists explore unseen chemical space and select high-yield reactions\cite{b5,b28}. We train our models with a relatively small training set (2.5\% of the Buchwald–Hartwig HTE data, 5\% of the Suzuki–Miyaura HTE data, 50\% of the Buchwald–Hartwig ELN data) and use them to predict the yields of the remaining reactions. The top 10 high-yield reactions are selected according to the prediction results. Then we calculate the average and standard deviation of 10 high-yield reactions. Fig. \ref{high-yield} presents the average and standard deviation of the yields for the top 10 reactions predicted to have the highest yields in the three datasets. Besides our method and baseline method, the result of ideal reaction selection and random reaction selection are presented. In all three datasets, our method has a higher average yield and lower standard deviation than baseline selection and random selection. For example, in the Buchwald–Hartwig HTE dataset, using MetaRF trained on 2.5\% of the dataset, the predicted top 10 high-yield reactions from the remaining dataset have an average yield of 93.7$\pm$4.1\%, compared to the ideal selection of 95.5$\pm$1.8\%. In contrast, baseline selection has an average yield of 86.3$\pm$5.4\% and random selection has an average yield of 52.1$\pm$24.9\%. The selection works similarly for the Buchwald–Hartwig ELN and Suzuki–Miyaura HTE dataset.

\subsection{Ablation Study}
To validate the effects of each component in MetaRF, we conduct an ablation study on the Buchwald–Hartwig HTE dataset, with 20\% of the data as the training set. The number of fine-tune samples is five in the ablation study. For the baseline method (random forest), five fine-tune samples are randomly selected and then added to the training set.

The first ablation replaces the dimension-reduction based sampling with random sampling. The random sampling experiment is repeated 10 times, and average performance is recorded. The second ablation removes the random forest structure, using MAML to replace the MetaRF framework. The third ablation keeps the random forest structure and uses a standard pretraining and fine-tuning framework in transfer learning\cite{b19} to replace MAML. 

Table \ref{ablation} presents the comparison results of predicting performance in terms of $R^{2}$ and RMSE. When dimension-reduction based sampling is replaced with random sampling, the $R^{2}$ decreases from 0.7738 to 0.7003, demonstrating the effectiveness of the dimension-reduction based sampling method. The results of the ablation study also clearly demonstrate the importance of random forest structure in MetaRF. Removing random forest causes $R^{2}$ performance to decrease from 0.7738 to 0.3730, which shows that random forest can tackle the overfitting problem in few-shot prediction. Regarding the results of the third ablation test, $R^{2}$ decreases by 10\% when MAML is replaced with transfer learning, and transfer learning has minor improvement compared to the baseline.

\begin{figure*}[htbp]
\centerline{\includegraphics[width=0.65\textwidth]{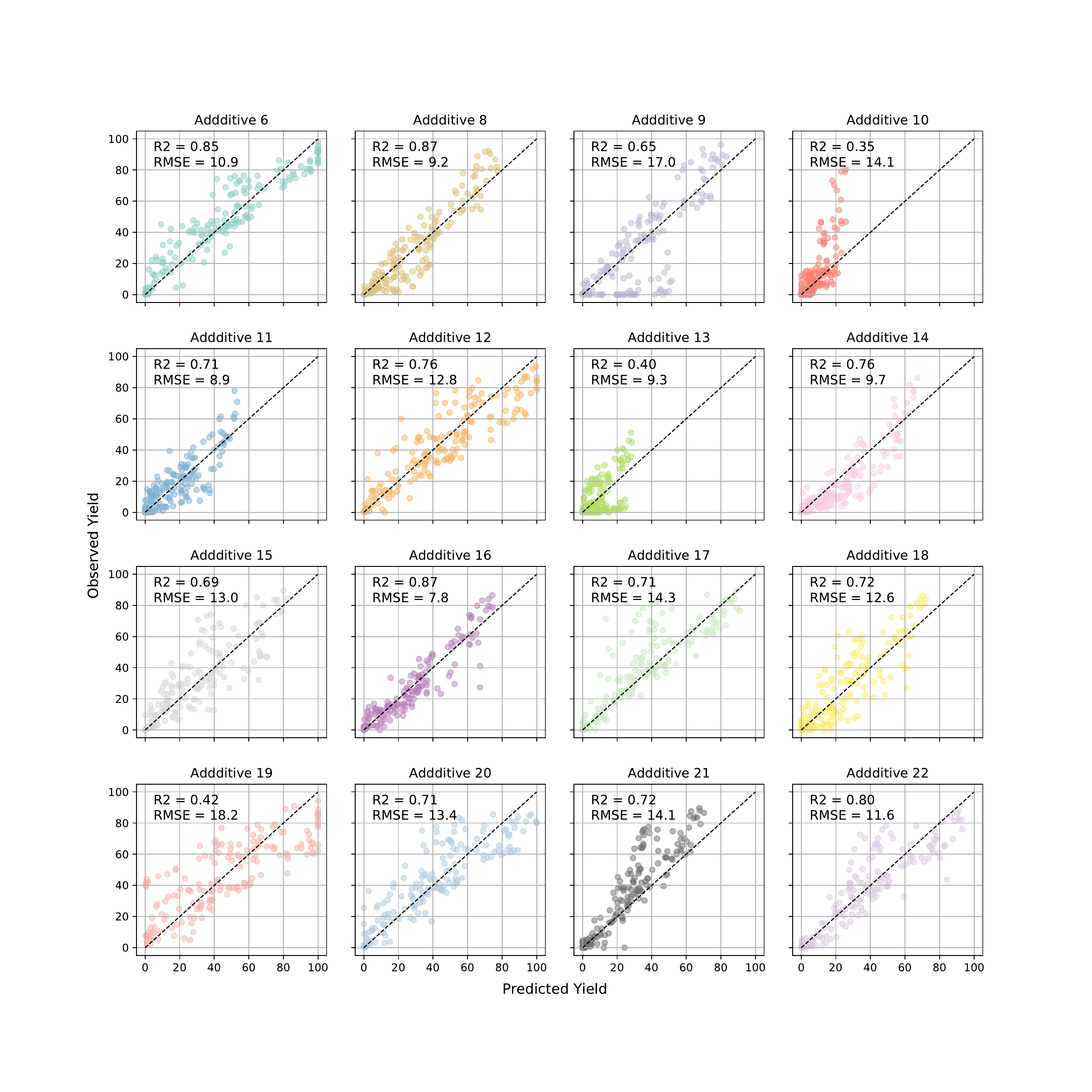}}
\caption{The prediction results on different additives. For each additive in the testing set, the predicted yield and observed yield is presented in a subplot. The title of each subplot is the index number of the additive.}
\label{scatter}
\end{figure*}

\begin{figure}[htbp]
\centerline{\includegraphics[width=0.5\textwidth]{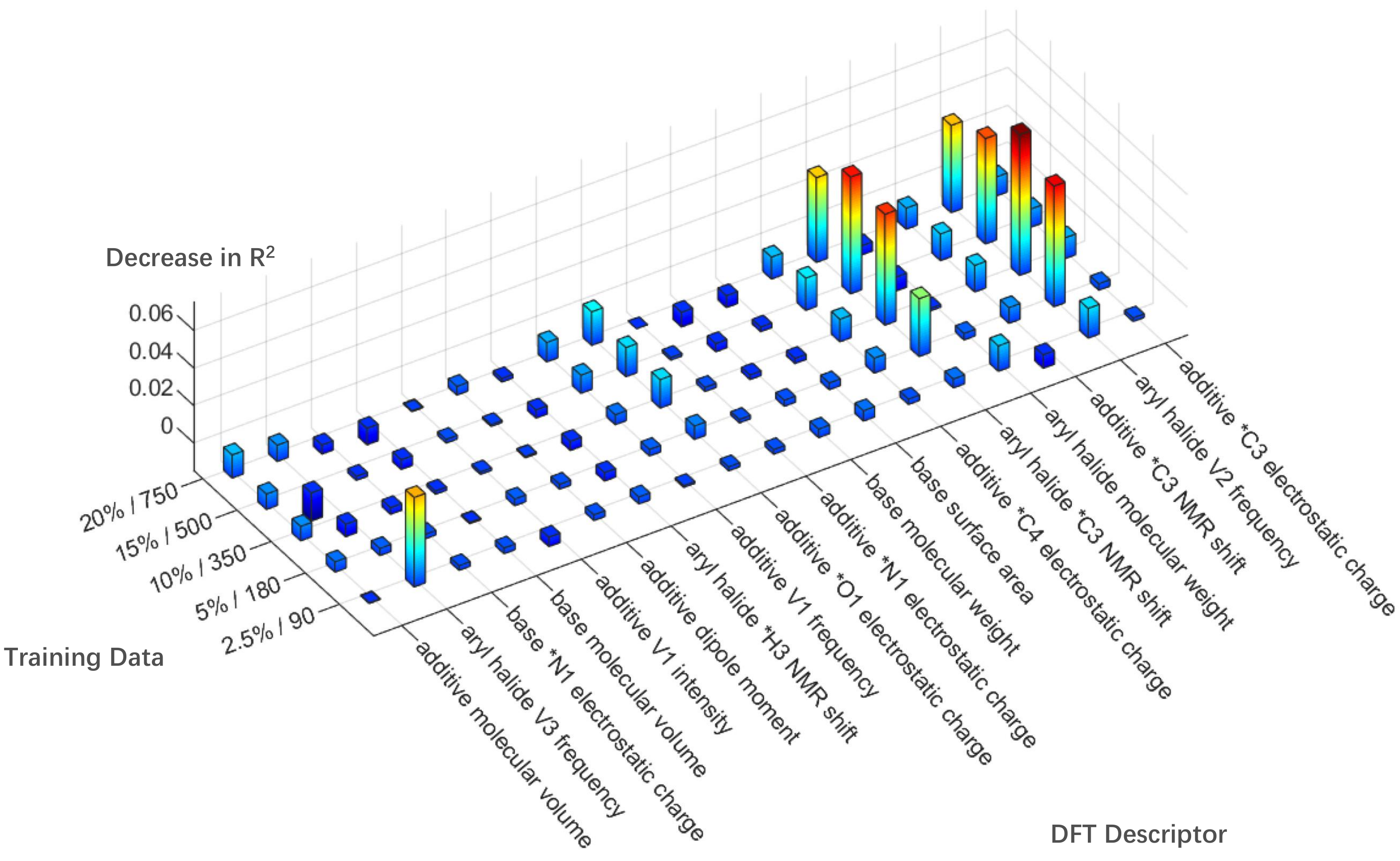}}
\caption{Most important DFT descriptors of the model trained on different sizes of data. Feature importance is determined by the decrease in $R^{2}$ upon reshuffling the values of the feature. * indicates a shared atom. E indicates energy; HOMO indicates the highest occupied molecular orbital; V indicates vibration.}
\label{result_feature_importance}
\end{figure}

\subsection{Analysis on Fine-tune Sample Number}
We analyze the effect of adjusting the number of fine-tune samples on the Buchwald–Hartwig HTE dataset\cite{b4}, using 20\% of the data as the training set. The few-shot yield predicting ability is tested by root mean square error (RMSE) and $R^{2}$ performance. Fig. \ref{finetune_sample} is the performance valuation of our method compared to the baseline, random forest model. We can see that when the fine-tune sample varies from 2 to 15, MetaRF is significantly superior to the baseline in RMSE and $R^{2}$ performance. 

When the number of fine-tune samples is 5, we obtain an 18.82\% relative improvement in the $R^{2}$ performance and an 19.25\% relative improvement in the RMSE performance. Our method reaches $R^{2}$=0.7738 and RMSE=12.6401, while the $R^{2}$ and RMSE of baseline method (random forest) is 0.6538 and 15.6535, respectively. The evaluation results of relative improvement are listed in Table \ref{margin}. When the number of fine-tune samples varies, the RMSE relative improvement is still around 20\%, which demonstrates the stable and satisfactory performance of MetaRF on a few-shot yield prediction.

\subsection{Predicting Performance on Each Additive}
For Buchwald–Hartwig HTE dataset, when using 20\% of the data as the training set, the predicting performance of each additive in the testing set is shown in Fig. \ref{scatter}. In this experiment, the number of fine-tune samples is 5. For each additive, the predicted yield and observed yield are presented in a subplot. From Fig. \ref{scatter}, we can see that our model has satisfactory performance on new additives in the testing set, which shows that our model can quickly adapt with only 5 data points.

\subsection{Interpretability Analysis}

For interpretability analysis, we visualize the most important DFT (Density Functional Theory) descriptors in the model trained on different sizes of Buchwald–Hartwig HTE data in Fig.\ref{result_feature_importance}. One measure of feature importance is the decrease in the model's $R^{2}$ performance when the values of that feature are randomly shuffled, and the model is retrained\cite{b4}. The feature importance results of models trained on different sizes of data have a slight difference. Generally, the most important descriptors are aryl halide’s *C3 nuclear magnetic resonance (NMR) shift (the asterisk indicates a shared atom), aryl halide’s vibration frequency, additive's *C3 NMR shift and additive's *C3, *O1, *C4 electrostatic charges.

~\\
\section{Conclusion}\label{sec13}
This paper proposes an attention-based random forest model to solve the few-shot yield prediction problem. The workflow includes using the DFT feature to encode chemical reactions and using the meta-learning framework to decide the attention weights of random forest. In the fine-tuning phase, we only need several samples to acquire satisfactory performance on new reagents. Our method obtains about 20\% lower RMSE when the fine-tune sample varies from 4 to 10. The effective few-shot  prediction demonstrates that our method can predict the effect of a new reactant structure with few additional data. The methodology in this paper brings benefits to future work on few-shot yield prediction.

\bibliographystyle{IEEEtran}
\bibliography{snbibliography}

\end{document}